\documentclass[conference]{IEEEtran}
\IEEEoverridecommandlockouts
\usepackage{cite}
\usepackage{amsmath,amssymb,amsfonts}
\usepackage{algorithmic}
\usepackage{graphicx}
\usepackage{textcomp}
\usepackage{xcolor}
\usepackage{comment}
\usepackage{caption}
\usepackage{enumitem}
\usepackage{multicol}
\usepackage{float}
\usepackage{bm}
\usepackage{algorithmic}
\usepackage{textcomp}
\usepackage{xcolor}
\usepackage{url}
\usepackage{amsmath,}
\usepackage{multirow}

\graphicspath{ {images/} }

\def\BibTeX{{\rm B\kern-.05em{\sc i\kern-.025em b}\kern-.08em
    T\kern-.1667em\lower.7ex\hbox{E}\kern-.125emX}}
\begin{document}

\title{\title{Contextual Networks and Unsupervised Ranking of Sentences}}

\author{\IEEEauthorblockN{Hao Zhang}
\IEEEauthorblockA{\textit{Dept. of Computer Science} \\
\textit{University of Massachusetts}\\
Lowell, USA \\
hao\_zhang@student.uml.edu}
\and
\IEEEauthorblockN{You Zhou}
\IEEEauthorblockA{\textit{Dept. of Computer Science} \\
\textit{University of Massachusetts}\\
Lowell, USA \\
you\_zhou@student.uml.edu}
\and
\IEEEauthorblockN{Jie Wang}
\IEEEauthorblockA{\textit{Dept. of Computer Science} \\
\textit{University of Massachusetts}\\
Lowell, USA \\
wang@cs.uml.edu}
}

\maketitle

\begin{abstract}
We construct a contextual network to represent a document with syntactic and semantic relations 
between word-sentence pairs, based on which 
we devise an unsupervised algorithm called CNATAR 
to
score sentences,
and rank them through a bi-objective 0-1 knapsack maximization problem 
over topic analysis and sentence scores.
We show that CNATAR outperforms the combined ranking of the three human judges provided on the SummBank dataset under both ROUGE and BLEU metrics, which in term significantly outperforms each individual judge's ranking. Moreover, CNATAR produces so far the highest ROUGE scores over DUC-02, and outperforms previous supervised algorithms on the CNN/DailyMail and NYT datasets. We also compare the performance of CNATAR and the latest supervised neural-network summarization models and compute oracle results.
\end{abstract}

\begin{IEEEkeywords}
contextual network, topic analysis, T5 sentence similarity, bi-objective 0-1 knapsack
\end{IEEEkeywords}

\section{Introduction} 

%
Ranking sentences (or segments of text) for a given article 
may be used, for example, as an oracle to build a hierarchical-reading tool to allow readers to read the article one layer of sentences at a time in a descending order of significance,
as a selection criterion to construct a better search engine, or as a base for constructing a
 summary.

We present an unsupervised algorithm called CNATAR (\textbf{C}ontextual \textbf{N}etwork \textbf{A}nd \textbf{T}opic \textbf{A}nalysis \textbf{R}ank) to rank sentences for a given article,  which works as follows: 

Step 1: Construct
 a contextual network (CN) to represent semantic and syntactic relations between sentences in the article 
by leveraging dependency trees and contextual embeddings of words
to form weighted edges
between word-sentence pairs. 
%

Step 2: Devise an unsupervised algorithm called CNR (Contextual Network Rank) to score 
nodes of the underlying CN 
using a biased PageRank algorithm w.r.t.
the underlying article structure,
and then score a sentence by summing up node scores for nodes
containing the said sentence with a BM25 normalizer. 

Step 3:
Carry out topic analysis using Affinity Propagation  \cite{dueck2009affinity} 
based on T5 sentence similarity, and rank sentences by approximating a bi-objective 0-1 knapsack maximization problem to select sentences with the largest scores and topic diversity using the Within-Cluster Sum of Square metric and
dynamic programming.

We show that CNATAR outperforms the combined ranking of all human judges over the SummBank dataset in all categories under both ROUGE and BLEU measures, 
and substantially outperforms each judge's individual ranking.
Moreover, CNATAR is efficient with an average running time of about 0.7 seconds for
each document in SummBank on a commonplace CPU desktop computer. 
We also evaluate CNATAR on other datasets for abstractive summaries, including DUC-02, CNN/DailyMail (CNN/DM in short), and NYT. We show that
CNATAR outperforms all previous algorithms on DUC-02; and 
outperforms all previous unsupervised algorithms and the supervised model REFRESH  \cite{narayan2018ranking} on CNN/DM and NYT 
trained on these datasets. We then compare performance of CNATAR and the two latest supervised BERT-based models BERTSum \cite{liu2019fine} and MatchSum \cite{zhong2020extractive}. 

\section{Related Work} \label{sec:related work}

Early sentence-ranking algorithms typically score sentences 
in connection to text summarization.
Recent unsupervised methods include $\text{CP}_3$ \cite{parveen2016generating},
Semantic SentenceRank (SSR) \cite{zhang2020unsupervised}, BES (BERT Extractive Summarizer) \cite{miller2019leveraging} and PacSum \cite{zheng2019sentence}.

$\text{CP}_3$  models a document as a bipartite graph between words and sentences and uses
Hyperlink-Induced Topic Search 
\cite{kleinberg1998authoritative} 
to score sentences 
that maximizes sentence importance, non-redundancy, and coherence.
%

SSR introduces semantic relations overlooked by early unsupervised algorithms 
to construct word-level and sentence-level semantic graphs. 
It uses article-structure-biased (ASB) PageRank to score words and sentences separately,
and then combines them to generate the final score for each sentence. SSR ranks sentences based on their final scores and topic diversity through semantic subtopic clustering. 
In so doing, SSR offers higher ROUGE scores on the DUC-02 dataset than CP$_3$ and
the previous unsupervised algorithms, and significantly outperforms each judge's individual ranking on the SummBank dataset, but still falls short of the
combined ranking of the three judges.
BES
clusters sentence embeddings
generated by BERT with K-means, and ranks a sentence by the Euclidean distance between the sentence and the centroid of the underlying cluster. PacSum builds a complete graph based on dot products of sentence embeddings, attempting to capture influence of 
any two sentences to their respective importance 
by their relative positions in the document. 

Recent supervised 
methods construct neural-network models 
to perform sequence scoring/labeling. REFRESH \cite{narayan2018ranking} 
a sentence with a CNN encoder and
scores sentences using LSTM  
that globally optimizes the ROUGE metric with reinforcement learning.
%
%
BERTSum \cite{liu2019fine}, a fine-tuned BERT embeddings for sentences
from the input document, 
scores sentences with a summarization-specific layer trained on a labeled dataset such as CNN/DM. 
MatchSum \cite{zhong2020extractive}, another variant of BERT, 
produces an embedding for the input document and an embedding of
the best summary candidate that is most similar to the document embedding. 
A summary candidate is formed with a desired number of sentences selected
from a number of sentences with high scores produced by other models such as BERTSum.
%
These supervised models are trained on the CNN/DM and NYT datasets, all imposing a small upper bound on the size of an input sequence due to the difficulty on handing a long sequence. 
BERTSum, for example, imposes an input sequence of upto 512 tokens (about 30 sentences on average) and drops the remaining text after the first 512 tokens.
%
%
%
%
Needing a large labeled dataset to train a supervised neural-network model also imposes a major roadblock for languages without such labeled datasets.

\section{Contextual Networks} \label{sec:cn}

\begin{figure}[h]
\centering
\includegraphics[width=0.49\textwidth]{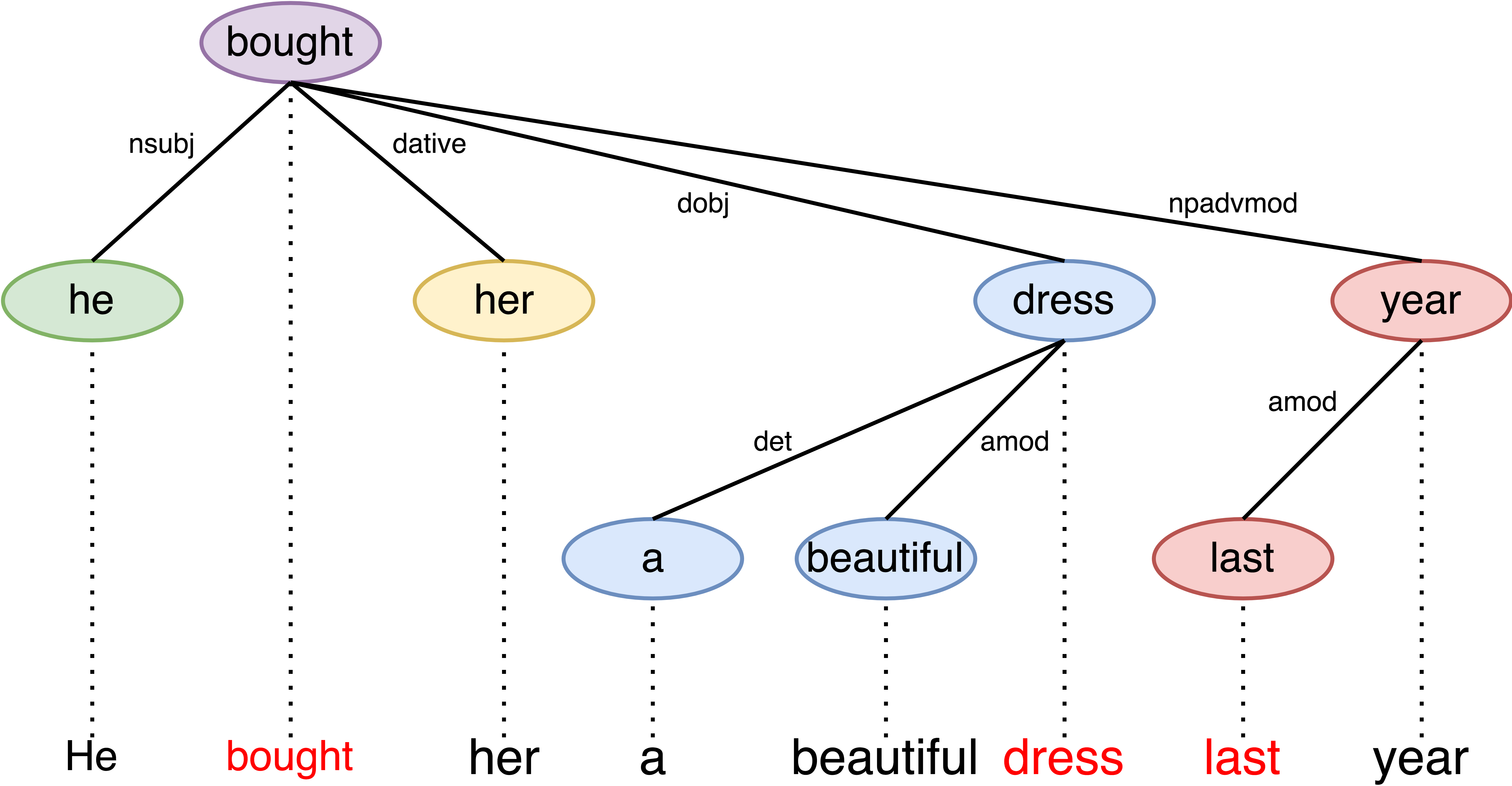}
\caption{A dependency tree for ``He bought her a beautiful dress last year."}
\label{fig:dep}
\end{figure}

Let $D$ denote a document of $m$ sentences and let $\langle S_1, S_2, \ldots, S_m\rangle$
be the original sequences of sentences in $D$.
Compute a dependency tree for each sentence and then
replace each pronoun in a sentence with its original mention
using
a coreference resolution tool.
A dependency tree for a sentence \cite{hudson1984word} is an undirected tree rooted at the main verb, connecting other words 
according to grammatical relations 
(see Fig. \ref{fig:dep} for an example). 
%
%
Compute a contextual embedding $e(w,S)$ for with $w \in S$.
Next, 
mark non-content words (stop words) using a stopword filter.
Stop words include determiners,  prepositions, postpositions, coordinating conjunctions, copulas, and auxiliary verbs.
%
For each sentence $S$, let $w$ and $w'$ be two content words in $S$. If there are 
stopwords $\sigma_1 \cdots, \sigma_r$ 
such that 
$w, \sigma_1, \ldots, \sigma_r, w'$ forms a path 
on the dependency tree $T_S$, 
then add a new connection of $w$ and $w'$ in $T_S$. 
%
Finally, remove stop words and replace every content word with its lemma using a lemmatizer. 

In what follows, unless otherwise stated, by 
``word" it means its lemma. 
%
For each word $w$ in $S$, let $N_w$ 
be the set of direct neighbors of $w$ on $T_S$.
Two words $x$ and $y$ are said to be
\textsl{syntactically related} if either they are neighbors (i.e., $x \in N_y$) or they share a common third neighbor $w$ (i.e., $x \in N_w$ and 
$y \in N_w$).
%
This relation captures 
the structure of 
subject-verb-object in the same sentence such that any two of
these words are syntactically related. 

Let $\langle w_1, w_2, \ldots, w_n\rangle$ be the original sequence of words in $D$.
By comparing $w_i$ with $w_j$ we mean to compare the words at locations $i$
and $j$. Denote by $(w_i, S_k)$ the $i$-th word in the $k$-th sentence.
When $i$ is given, it is straightforward to determine $k$, which can be 
expressed with a function $h(i)$. Namely, $w_i \in S_{h(i)}$ for all $i$. 
If $i \not= j$, 
then
$(w_i, S_{h(i)})$ and $(w_j, S_{h(j)})$ are different entities 
even if $w_i = w_j$ and $h(i)=h(j)$.

\begin{figure}[h]
\centering
\includegraphics[width=0.3\textwidth]{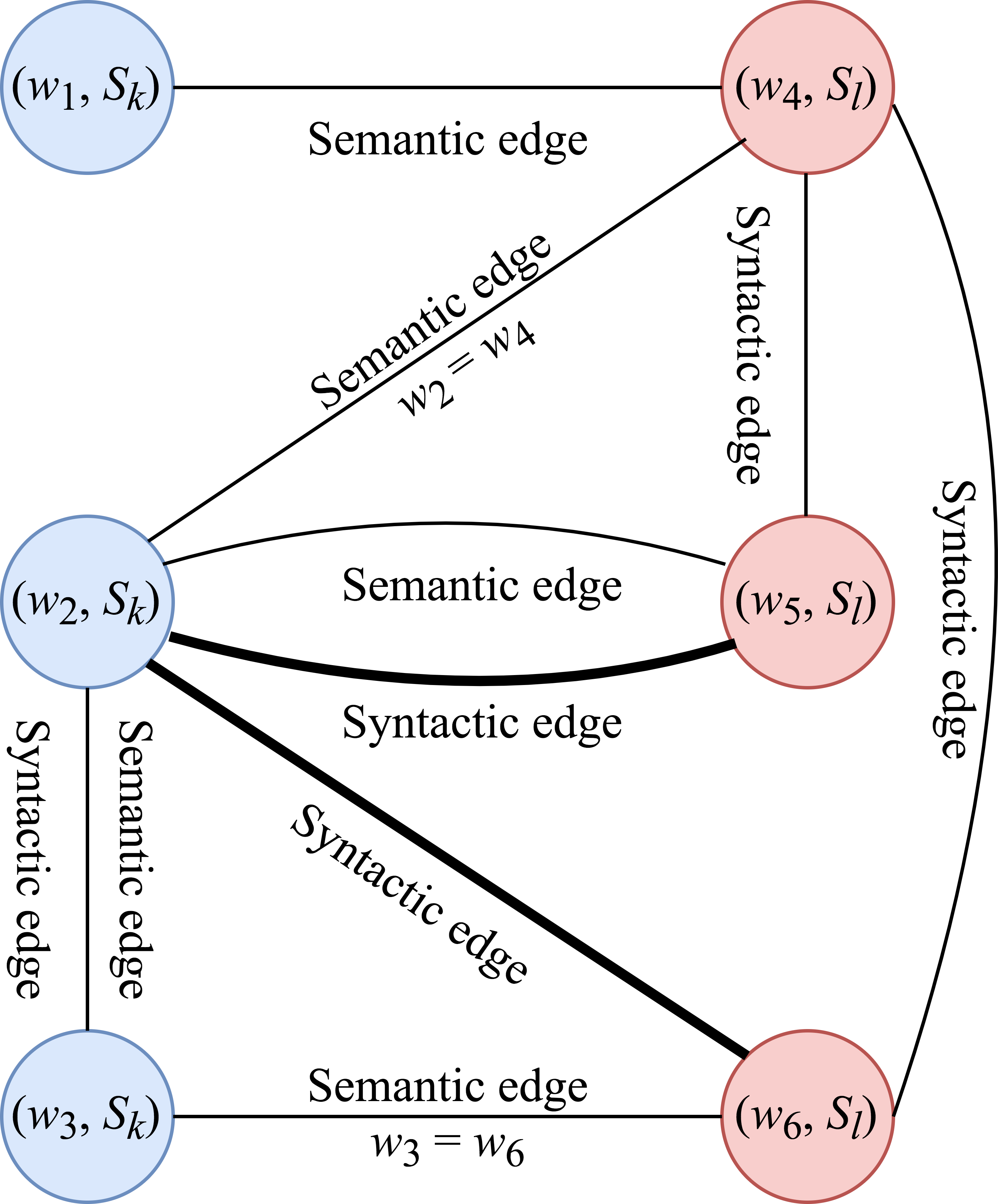}
\caption{
Two syntactic edges between $S_k$ and $S_l$.}
\label{fig:cn-init}
\end{figure}

Construct a weighted, undirected multi-edge graph $G_D = (V_D, E_D)$
with 
 $V_D = \{v_i \mid v_i = (w_i,S_{h(i)}), 1\leq i \leq n\}.$ 
%
%
Let 
$v_i,v_j \in V_D$ with $i \not =j$.
$E_D$ is constructed below:
%
%
(1) \textsl{Semantic edges inside or across sentences}. 
Connect $v_i$ and $v_j$ if the cosine similarity of $e(v_i)$ and $e(v_j)$ is
at least
$\delta$ (a hyperparameter;
it is reasonable to set $\delta= 0.7$).
%
(2) \textsl{Syntactic edges inside the same sentence}, namely, $h(i)=h(j)$.
Connect $v_i$ and $v_j$ if $w_i$ and $w_j$ are syntactically related on the dependency tree
$T_{S_{h(i)}}$.
%
(3) \textsl{Syntactic edges across sentences}, namely,
$h(i)\not= h(j)$. Connect $v_i$ and $v_j$
if there is a third node $v_{q} = (w_{q}, S_{h(q)})$ 
with $h(q) = h(i)$ and 
$q \not= i$
such that 
%
%
$w_{q} = w_j$, 
$v_q$ and $v_j$ are semantically connected as in (1), and
$v_q$ and $v_i$
are syntactically connected as in (2);
or the mirror condition (i.e., swap $i$ with $j$ in the above condition) is true.
Fig. \ref{fig:cn-init} illustrates this construction.
In other words, we transform a syntactic relation between $w_{q}$ and $w_i$ inside $S_{h(i)}$ 
to a syntactic relation between $v_i$ and $v_j$ 
if $w_q$ w.r.t. $S_{h(i)}$ is semantically close to $w_j$ w.r.t. $S_{h(j)}$. 
Note that requiring $w_{q} = w_j$ is critical, for it will result in undesirable syntactic relations 
if $w_q \not= w_j$
(see Remark 1 below).


To construct a contextual network for $D$,
compute edge weights and merge multiple edges.
If $v_i$ and $v_j$ are connected by a syntactic edge, 
let its initial weight be 1, and normalize it by the total number of syntactic edges.
If they are connected by a semantic edge, let its initial weight be
the cosine similarity of $e(v_i)$ and $e(v_j)$, and 
normalize it by the summation of all the initial weights of the semantic edges.
%
If $v_i$ and $v_j$ are connected by both a syntactic edge and a semantic edge, then
merge the two edges to one edge and let its new weight be the summation of the corresponding
syntactic weight and the semantic weight. 

%

\begin{figure}[h]
\centering
\includegraphics[width=0.4\textwidth]{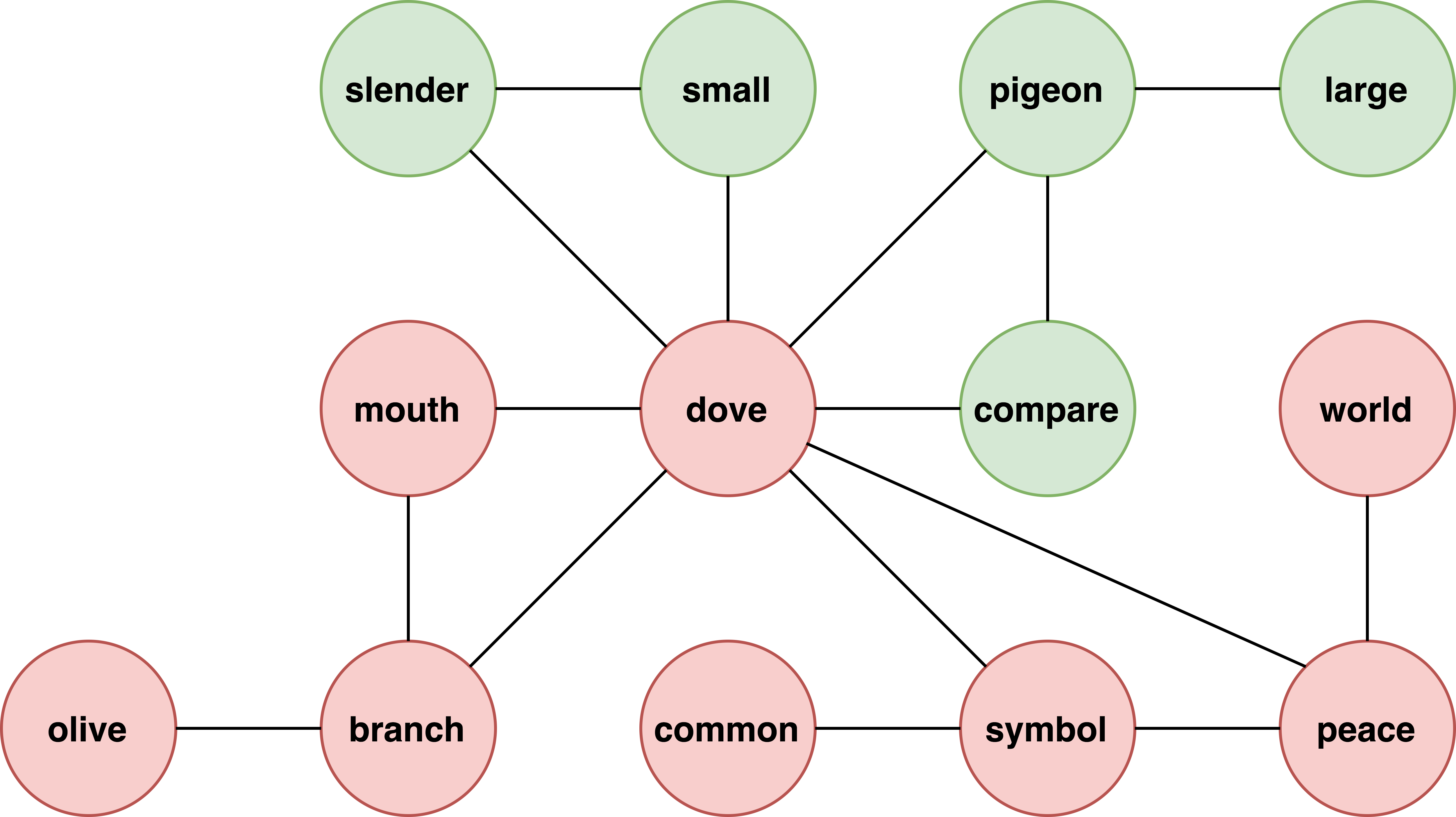}
\caption{Syntactic relations via dependency trees on
$S_1$ (red) and $S_2$ (green) mentioned in Remark 1.
}
\label{fig:cn-syntactic}
\end{figure}

\textbf{Remark 1}.
To see why we must require $w_{q} = w_j$ when constructing syntactic edges across sentences 
consider, for example, the following two sentences: 
$S_1$: \textsl{A dove with an olive branch in its mouth is a common symbol of world peace.}
$S_2$: \textsl{Doves, comparing with pigeons, are smaller and slenderer, while pigeons are larger.}
Fig. \ref{fig:cn-syntactic} depicts the correct syntactic relations by our construction. 
If, however, we allowed $w_{q} \not= w_j$, then
because the cosine similarity of $e(``dove",S_1)$ and
$e(``pigeion",S_2)$ is greater than the threshold value of $\delta = 0.7$,  
these two nodes 
would be connected by a semantic edge, implying that
``dove" in $S_1$ and ``large" in $S_2$
would be syntactically connected, which 
is undesirable.

\begin{figure}[h]
\centering
\includegraphics[width=0.4\textwidth]{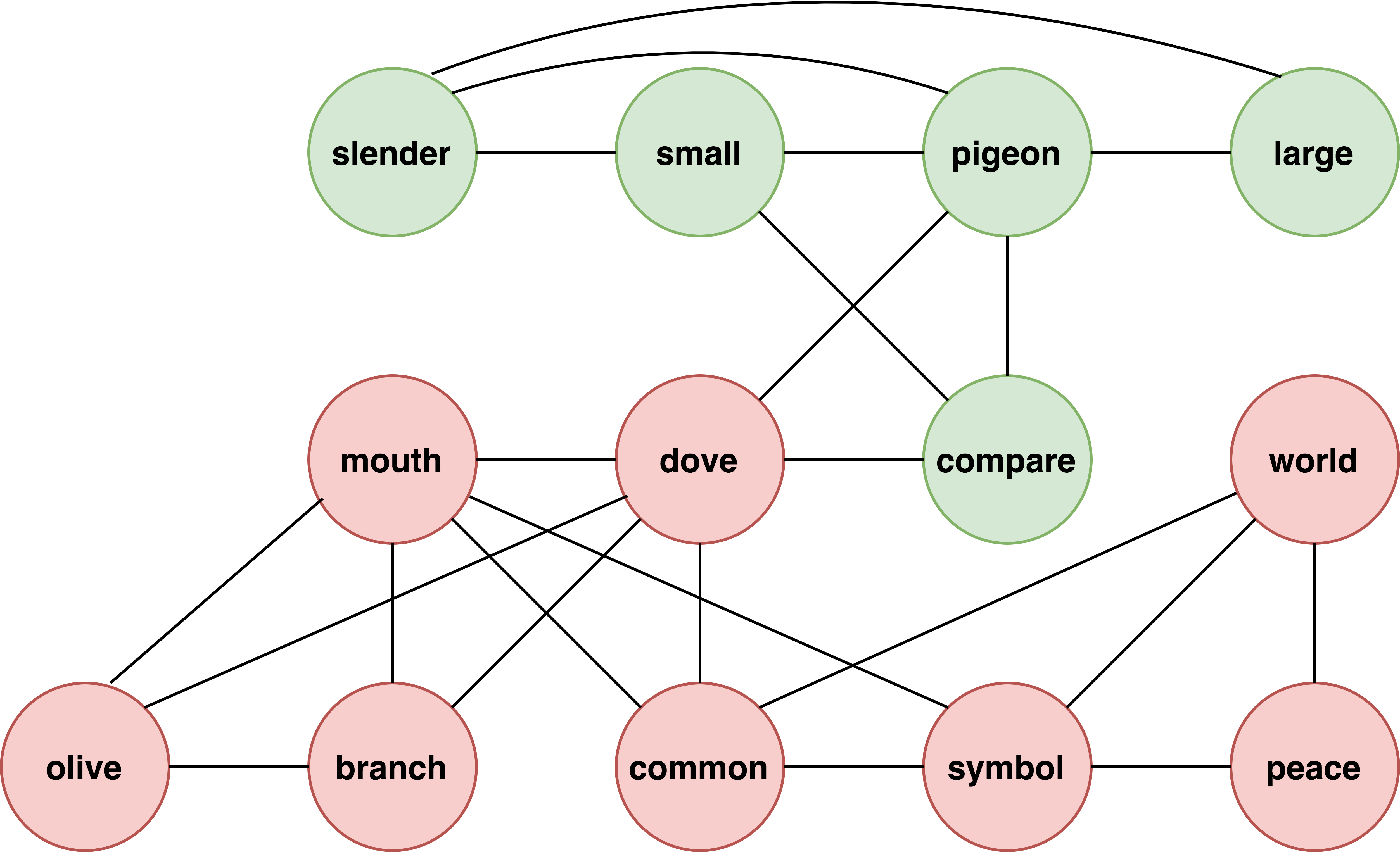}
\caption{Syntactic relations through co-occurrences with a window size of 3 between words in $S_1$ and
$S_2$ in Fig. \ref{fig:cn-syntactic}.}
\label{fig:cn-cooccur}
\end{figure}

\textbf{Remark 2}.
Co-occurrences of words are previously used to capture syntactic relations between words, where
two words are related if they co-occur in a small window of successive words. However, this method may falsely
relate unrelated words and miss related words.
For example, if two adjacent words in the same sentence fall in different sub-trees of its dependency tree, then they are unrelated from the
syntactic point of view, but they could be made related because they co-occur. Co-occurrence also fails to capture related words that do not co-occur within a small window. 
Fig. \ref{fig:cn-cooccur} depicts the syntactic relations of words in
the above sentences $S_1$ and $S_2$ with a window size of 3,
which includes undesirable syntactic edges between ``pigeon" and ``small",
``pigeon" and ``slender", and ``mouth" and ``symbol"; yet misses desirable syntactic edges between
``dove" and ``small", ``dove" and ``slender", ``dove" and ``peace", among other things.
Our construction of syntactically related words through dependency trees resolve these issues.

\section{Sentence Ranking}  \label{sec:scoring}

Article structures also play a role in ranking sentences \cite{zhang2020unsupervised},
which may be classified into four types based on locations where words tend to be more important: 
(1) \textsl{Rectangle}. Words are of the same importance in any part of the article.
Narrative articles are typically of this type. 
(2) \textsl{Inverted pyramid}.
Words toward the beginning of the article tend to be more important.
News articles are typically of this type. 
(3) \textsl{Pyramid}. Words toward the end of the article tend to be more important.
Argumentative articles are typically of this type. 
(4) \textsl{Hourglass}. Words toward the beginning and the end of the article tend to be more important. Research  papers are typically of this type. 
%

Let $\text{LW}(i) > 0$ denote the location weight of
the $i$-th word (to be constructed later) 
with $\sum_{i=1}^n \text{LW}(i) = 1$.
CNR computes the score of node $v_i$ over the contextual network, denoted by $\text{score}(v_i)$, using
the following article-structure-biased (ASB) PageRank: 
\begin{align*}
\text{score}(v_i) &= 0.85 M(v_i) + 0.15 \text{LW}(i), \text{ where}\\
M(v_i) &= \sum_{v_j \in Adj(v_i)} \frac{wt(v_i, v_j)\cdot \text{score}(v_j)}{\sum_{v_k \in Adj(v_j)} wt(v_j, v_k)},
\end{align*}
and $wt(u,v)$ is the edge weight of $(u,v)$.
It then
scores a sentence $S_k$ by summing up the scores of all the nodes $v_i$ with
$h(i) = k$ 
and normalizing the sum by a BM25 normalizer:
\begin{align*}
\text{score}(S_k) = \frac{\sum_{i: h(i)=k} \text{score}(v_i)}{1 - \beta + \beta\left(\frac{|S_k|}{\text{avsl}}\right)},
\end{align*}
where 
$|S_k|$ 
is the number of words
contained in $S_k$, 
$\text{avsl} = \sum_{j =1}^m |S_j|/m$ is the average sentence length
of the document, and $\beta \in [0,1]$ is a hyperparameter for the purpose of 
penalizing sentences that are longer than average  and rewarding sentences that are shorter than average.
Since the ratio of a sentence length over the average sentence length for a given document
is often larger than 2 or smaller than 1/2,
an appropriate value of $\beta$ should be near the first quadrant and we choose $\beta = 0.2$.

Next, we define $\text{LW}(i)$ 
so that it
does not abruptly change weight from location $i$ to location $i+1$. For the rectangle structure we simply use a uniform distribution with $\text{LW}(i) = 1/n$.
For the inverted pyramid structure, 
we use a slow decreasing quadratic curve to assign location weight for the $i$-th word by  
\begin{align*}
\text{LW}(i) =  \frac{6(\gamma-1)(i-n)^2}{(n-1)n(2n\gamma-n-\gamma)}+\frac{a(n-1)^2}{\gamma-1}, 
\end{align*}
where $\gamma = \text{LW}(1)/\text{LW}(n)$ is a hyperparameter (e.g., let $\gamma = 5$). 
The pyramid structure is a mirror image of the inverted pyramid, where the location weight of the $i$-th word equals the weight for the $(n-i+1)$-th word in the inverted pyramid structure.
For the hourglass structure, we again use a quadratic curve defined by
$\text{LW}(i) 
 = \left((i-n/2)^2+1\right)/\sum_{i=1}^{n}\left((i-n/2)^2 +1\right)$,
with the minimum value in the middle of 1 and $n$.
   

Sentence ranking 
should reflect 
the topics covered by the article.
A topic clustering algorithm partitions sentences 
into topic clusters based on a sentence similarity measure. 
Let $F(D')$ denote 
the 
distribution of topic covered by
a subset $D' \subseteq D$, and $L$ the maximum $|D'|$ allowed.
%
The sentence ranking problem can be modeled as the following bi-objective 0-1 knapsack maximization problem:
\begin{align*}
&\text{Maximize} \sum_{k=1}^m \text{score}(S_k)\cdot x_k \mbox{ and } F(\{S_k\mid x_k = 1\}), \\
&
\text{subject to}  \sum\limits_{k=1}^m  x_k = L \mbox{ and }x_k \in \{0, 1\},
\end{align*}
where 
$x_k = 1$ if $S_k$
is selected, and 0 otherwise. A ranking of sentences can be achieved by starting $L$ from 1, 
incremented by 1 each time, 
until $L = |D|-1$.

CNATAR approximates the bi-objective 0-1 knapsack problem as follows:
Suppose that $D$ is partitioned into $K$ topic clusters of sentences 
$D_1, \ldots, D_K$. 
Define a topic diversity function $F$ by 
dividing $L$ into $K$ numbers
$L_j = \lfloor(W_j /\sum_{\ell=1}^KW_\ell)L\rfloor,$ 
where 
$W_j$ is the Within-Cluster Sum of Square (WCSS) \cite{hartigan1979ak} for cluster $D_j$, which is the squared average distance of all the points within $D_j$ to the cluster centroid. Thus, $\sum_{j=1}^K L_j \leq L$.
Divide the bi-objective 0-1 knapsack into
$K$ 0-1 knapsack problems over each $D_j$ with
length bound $L_j$ for $1 \leq j \leq K$. That is, 
$\text{maximize }  \sum_{S_k \in D_j} \text{score}(S_k)\cdot x_k$, 
$\text{subject to} \sum_{S_k\in D_j} x_k = L_j \mbox{ and }x_k \in \{0, 1\}$,
where the $L_j$ sentences in $D_j$ with the highest scores form the maximum solution. 
Rank sentences in all solutions according to their scores. Let $L' = \sum_{j =1}^K L_j$. If $L' < L$, then select a remaining sentence with the highest score,
rank it after the selected sentences, and increase $L'$ by 1. 
Repeat until $L' = L$.

\textbf{Remark 3}. Sometimes we may need to select sentences such that the total number of words contained in them do not exceed a certain limit $L'_j$. The constraint of the 0-1 knapsack 
becomes $\sum_{S_k \in D_j} x_k\cdot |S_k| \leq L'_j$.
Using dynamic programming we can obtain a maximum solution to this version of
the $j$-th 0-1 Knapsack problem in $O(|D_j|L'_j)$ time, which
is feasible in practice since $|D_j|$ and $L'_j$ would be small.
We will need to use this version of 0-1 knapsack later when we deal with the DUC-02 dataset.

\section{Implementation and Evaluation} \label{sec:evaluation}

\begin{figure}[h]
\centering
\includegraphics[width=\columnwidth]{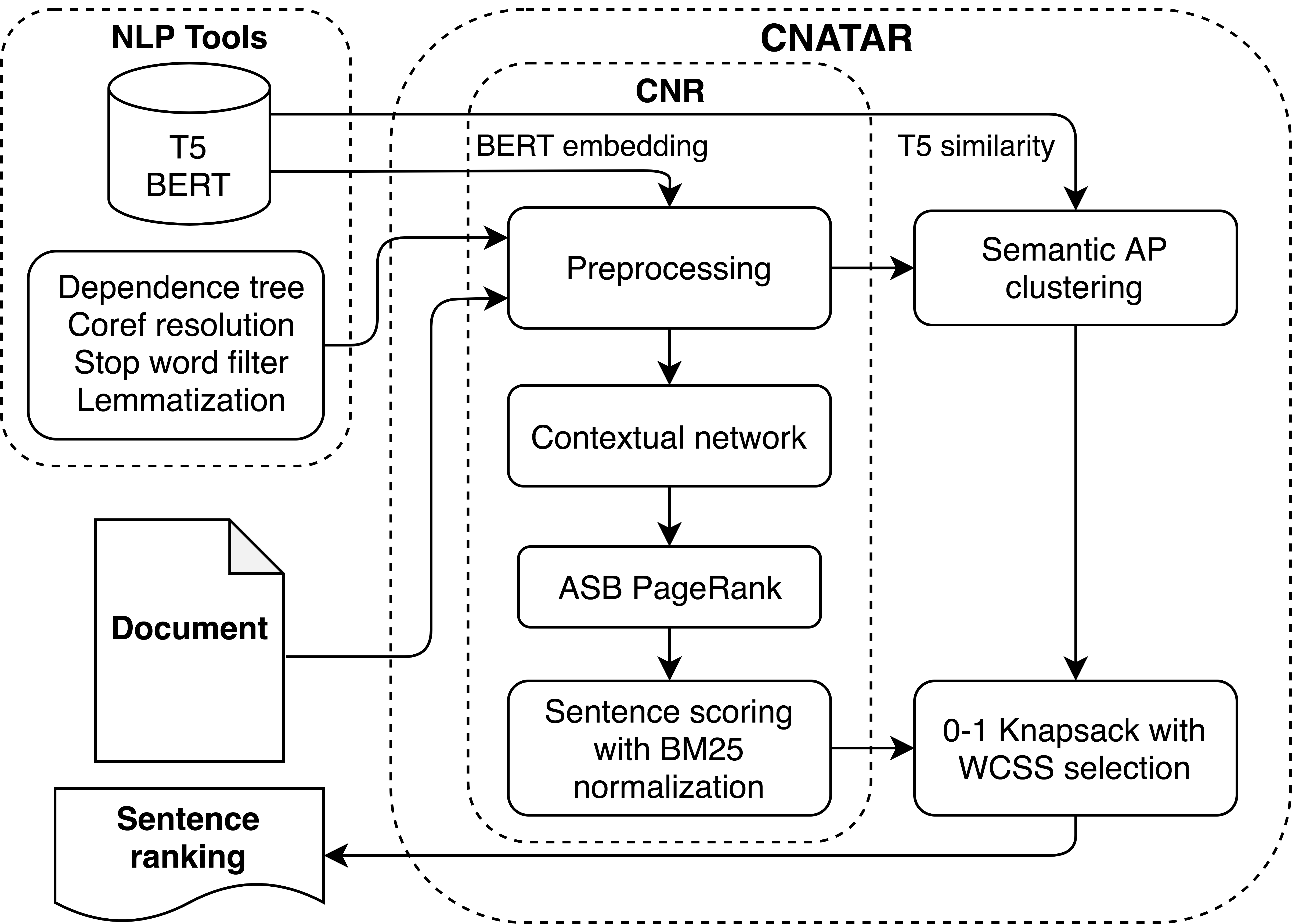}
\caption{CNATAR components and dataflows}
\label{fig:scheme}
\end{figure}

%
We preprocess documents with spaCy \cite{honnibal2017spacy} to split the text into sentences, and resolve coreference within a sentence using the NeuralCoref pipline \cite{wolf2017state}.
We then generate, for each sentence,
a dependency tree with spaCy, and generate contextual embedding using BERT-Large 
 \cite{devlin2018bert} 
for each word in the sentence. Next, we identify stopwords with spaCy's
stopword list 
and replace each content word with its lemma using spaCy's lemmatizer. 
To generate a contextual embedding for each word w.r.t. to a sentence, we sum up, 
the corresponding vector representations in the
last 4 layers of BERT-Large to form a contextual embedding of the word, 
to take the advantage of more syntactic information at the lower layers 
more semantic information at the higher layers  \cite{jawahar2019does}.
Finally, 
we use Affinity Propagation (AP) \cite{dueck2009affinity},  an exemplar-based clustering algorithm, to cluster sentences using a pretrained T5 similarity \cite{raffel2019exploring}
to compute sentence similarities. T5 similarity takes two sentences as input and returns a similarity score between 1 and 5.
AP dynamically determines the number of topic clusters. 
The major components and dataflows of CNATAR are shown in Fig. \ref{fig:scheme}.

\begin{table*}[ht]
\centering
\caption{Comparison of CMB-HR, CNATAR, CNR, SSR, PacSum, and BES against all judges over SummBank}
\label{table:1}
 \includegraphics[width=\textwidth]{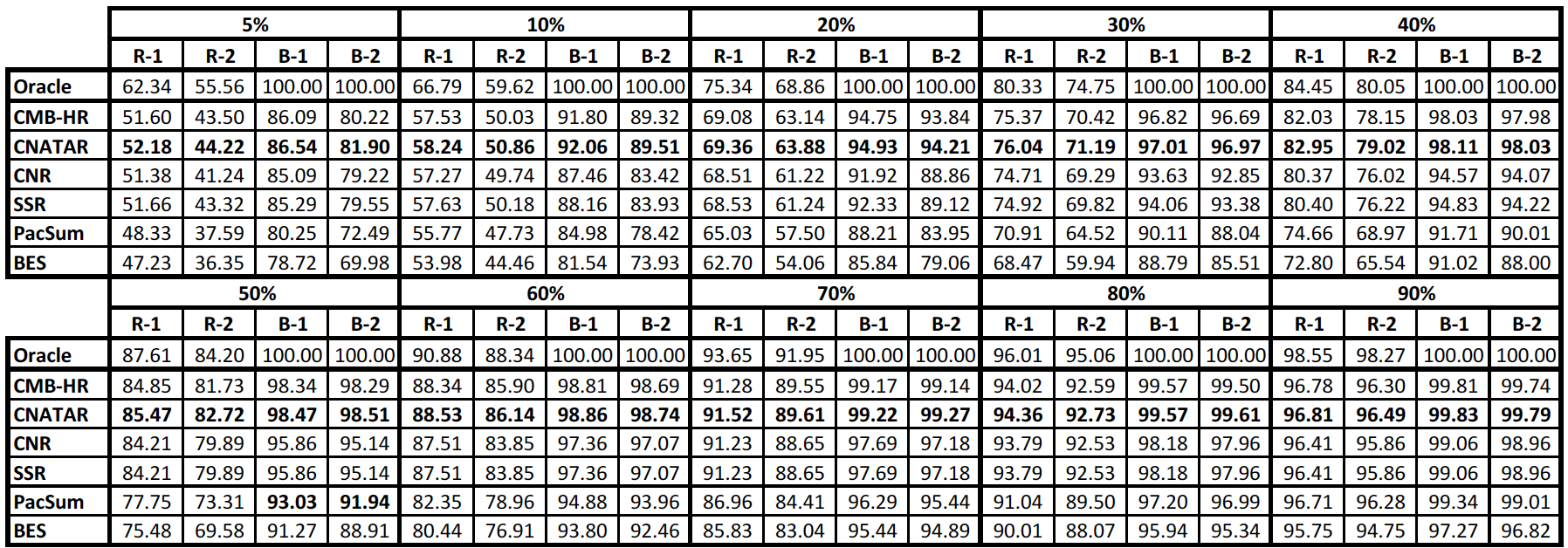}
\end{table*}

\textbf{Datasets}.
SummBank \cite{radev2003summbank} is the most suitable dataset for evaluating sentence-ranking algorithms. 
Three human judges rank sentences for each of the 200 news articles written in English individually in categories of top 
5\%, and 10\% to 90\% with an increment of 10\%. A combined sentence ranking of all judges, denoted by CMB-HR, is also provided on each article. 
%
%
%
%
%
%
DUC-02, CNN/DM, and NYT are other datasets for evaluating single-document summarization
algorithms, 
consisting of one or more human-written abstractive summaries for each article as the gold standard. 
Each summary in DUC-02 consists of upto 100 words, 
while each summary consists of an average of 3 sentences in CNN,
and 4 sentences in DM and NYT.
%
These datasets, although not ideal for evaluating sentence ranking, are used to 
compare with the latest summarization algorithms in the last 5 years. We follow the standard split
of training and evaluating \cite{nallapati2016abstractive} of CNN/DM on supervised algorithms, and use scripts supplied by \cite{see2017get} to obtain non-anonymized version of data. 
%
The XSum \cite{narayan2018don} dataset 
provides a one-sentence abstractive summary for each article, and so is inappropriate for evaluating sentence-ranking algorithms.

All of these datasets are news articles and so the location weight function for the inverted pyramid structure is applied.

\textbf{Comparison on SummBank}. 
We compare machine rankings with CMB-HR against each individual ranking as reference
and average the ROUGE \cite{lin2004rouge} and BLUE  \cite{papineni2002bleu} scores over all documents.
Both
CMB-HR and SSR outperform each individual judge's ranking using the other two judges' ranking as reference \cite{zhang2020unsupervised}. 
%
%
A full-range comparison is shown in Table \ref{table:1} against all judges under common measures of ROUGE-$n$ (R-$n$)  
and BLEU-$n$ (B-$n$), where $n = 1, 2$. The highest score under each category is shown in boldface. It can be seen that under all categories, CNATAR outperforms CMB-HR and substantially outperforms SSR, PacSum, and BES. SSR slightly outperforms CNR.
%
The oracle results are computed by choosing, for each article and under each percentage category,
an individual judge's selection of sentences that has the highest R-1 score against all three judge's selections. Because one judge's selection is always selected, the corresponding BLEU score is 100\%. 
%
We carry out the same experiments on two 32G-RAM computers, one with an Intel Core i7-8700K CPU
and the other an NVIDIA RTX 2080 Ti GPU. The average running time of CNATAR on each document 
is 0.73 seconds on the CPU machine, and 
0.6 seconds on the GPU machine.

\textbf{Comparison on DUC-02}.
Table \ref{tab:1} depicts the comparison results of the algorithms published in the last five years on the DUC-02 dataset, where each of the algorithms  
extracts sentences of the highest ranks with a total length bounded by 100 words.
Among these algorithms, CNN-W2V \cite{zhang2016extractive} is a supervised algorithm.
%
%
In addition, we also provide oracle results  by selecting a subset of sentences for each
document that maximizes the ROUGE score w.r.t. the benchmark summaries 
except the 6 articles with 78 sentences or more. 
For these 6 articles we use an approximation to avoid combinatorial explosion by selecting the first sentence with the highest R-1 score, then the next to the already-selected sentences with the highest R-1 score until the total number of
words exceeds 100. To the best of our  knowledge, no oracle results on DUC-02 were published before.
%
\begin{table}[h]
\centering
\caption{Comparison results (\%) on DUC-02, where the italic numbers 
are 
extracted from the corresponding papers.}
\label{tab:1}
\begin{small}
\begin{tabular}{l|c|c|c}
\hline \bf Methods & \bf ~R-1~ & \bf ~R-2~ & \bf ~R-SU4 \\ \hline
Oracle &52.0& 29.1 & 29.2\\
\hline
CNATAR 				& \bf 49.4 	& \bf 25.6 	& \bf 26.7 \\ 
CNR &49.2 & 24.8 & 26.1\\
SSR 				&  \textsl{49.3} 	&  \textsl{25.1} 	&  \textsl{26.5}   \\
$\text{CP}_3$				& \textsl{49.0}			& \textsl{24.7}		& \textsl{25.8} \\
PacSum    & \textsl{48.7}          & \textsl{23.3}     & \textsl{25.3} \\
CNN-W2V 	& \textsl{48.6}			& \textsl{22.0}		& $-$	\\
BES 				&  \textsl{48.5} 	&  \textsl{23.3} 	& \textsl{25.4}     \\
\hline
\end{tabular}
\end{small}
\end{table}
It can be seen that CNATAR outperforms all previous algorithms, supervised
and unsupervised. The three latest supervised models
trained on CNN/DM and NYT only produce 3 to 4 sentences for a given document, and so perform poorly on DUC-02, where each summary typically contains more than 4 sentences.


\textbf{Comparison on CNN/DM and NYT}.
Table \ref{tab:4} shows the comparison results of CNATAR, REFRESH \cite{narayan2018ranking}, 
BERTSum \cite{liu2019fine}, and MatchSum \cite{zhong2020extractive} on CNN/DM, NYT, and 
SummBank-4, a subset of 156 articles in SummBank that provide the top 4 sentences (the rest
of the articles do not provide top 4 sentences because SummBank only rank sentences on certain percentages).
All models output 4 sentences.
%
%
Recall that MatchSum suffers from a combinatorial blowup, to make it feasible to train,
we select sentence candidates using the top 5 most important sentences on CNN/DM, top 6 sentences on NYT, and top 9 sentences on SummBank-4.
%
It can be seen that CNATAR outperforms the unsupervised PacSum and the supervised REFRESH while MatchSum achieves the highest ROUGE scores, where R-L stands for ROUGE-L. On SummBank-4, CNATAR outperforms all the supervised models by a large margin, even if MatchSum has tried all 
possible candidate outputs for each
article in SummBank-4. The oracle results are computed 
by selecting the first sentence with the highest R-1 score, then select the next sentence to the already selected sentences with the highest R-1 score. 

\begin{table}[h]
\centering
\caption{Comparison results (\%), where the numbers in italic are taken from the corresponding papers.}
\label{tab:4}
\includegraphics[width=\columnwidth]{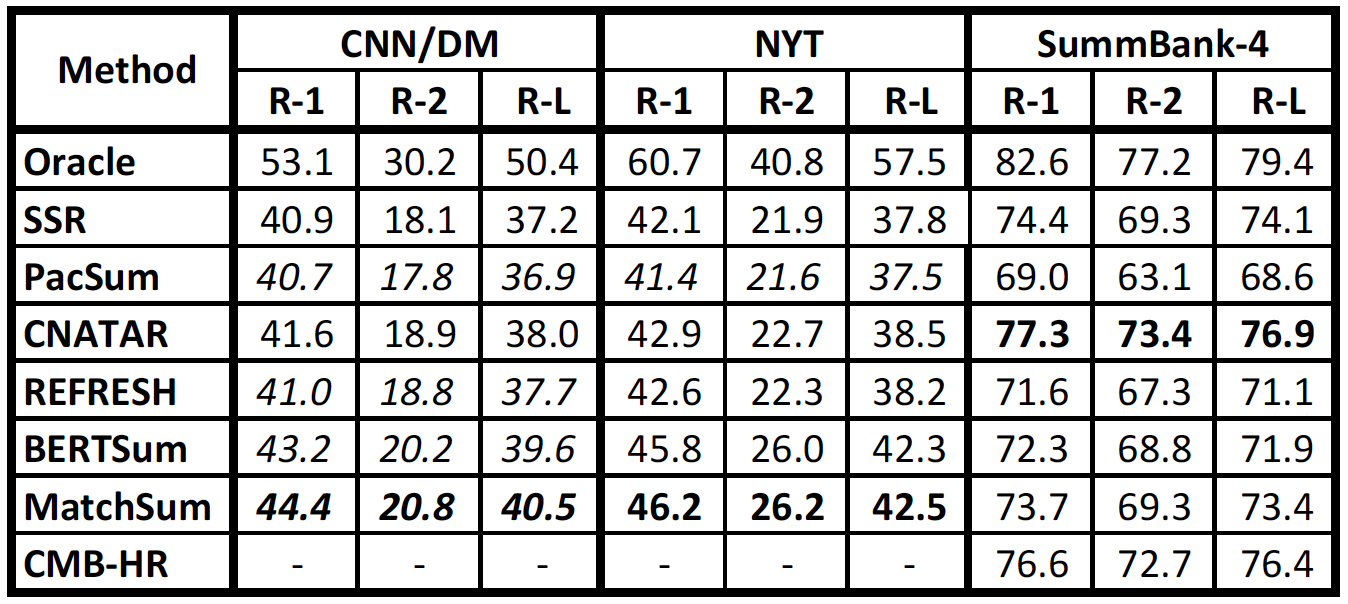}
\end{table}
\textbf{Ablation study}.
We show that, over SummBank, each mechanism in CNATAR is necessary for achieving its overall performance. In particular, contextual networks, location weights, and topic-cluster-wise 0-1 knapsack are the most significant components. Table \ref{tab:5} 
depicts the numerical results, where
\begin{table}[h]
\centering
\caption{Results from ablation study.}
\label{tab:5}
\includegraphics[width=\columnwidth]{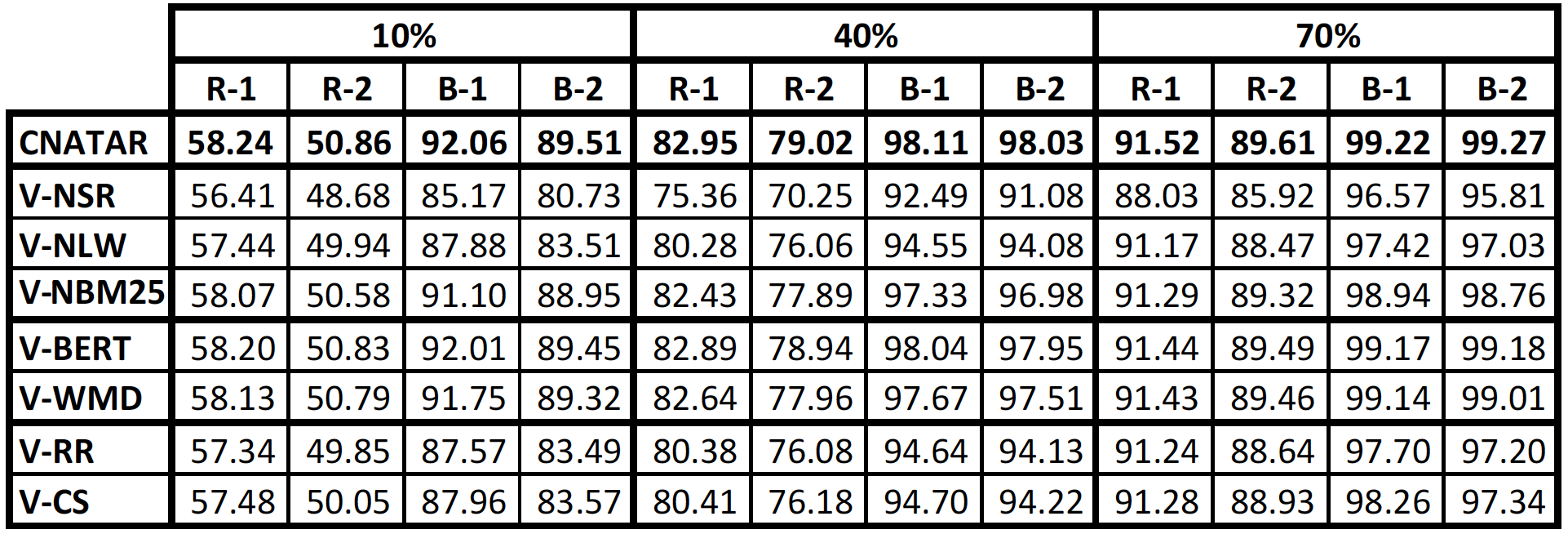}
\end{table}
%
V-NSR denotes a variant of CNATAR without contextual networks but using co-occurrences to capture weaker syntactic relations between words as in SSR 
\cite{zhang2020unsupervised}, V-NLW denotes a variant without location weight functions, and V-NBM25 a variant that replaces
the use of a BM25 normalizer with the standard normalizer of
sentence length. Moreover, V-BERT and V-WMD denote two variants that replace the T5 similarities with, respectively,  the
cosine similarity of BERT embedding, and similarities based on
Word Mover's Distance \cite{kusner2015word} as in SSR. 
Finally, V-RR and V-CS denote two variants that replace
the cluster-wise 0-1 knapsack with,  respectively, round-robin selections from clusters as in SSR and 
proportional selections based on cluster size.
%


\section{Conclusions and Final Remarks} \label{sec:conclusion}

CNATAR ranks sentences based on context networks and topic analysis, and achieves the state-of-the-art results. 
Our construction of contextual networks, however, only takes advantage of a few recent NLP tools. More NLP tools may be leveraged, including part-of-speech tags, role labeling, and sentiment analysis. Using these extra language features, it is expected 
a more appropriate weight can be computed when merging two edges in constructing a contextual network,  instead of assigning an equal weight as in the current
construction. 
%
Topic diversity also plays an important role in ranking sentences, and so it would be interesting to investigate a better mathematical formulation for the diversity function and explore other topic clustering algorithms. 

\bibliographystyle{IEEEtran}
\bibliography{anthology}

\end{document}